\documentclass{article}

 \usepackage[dblblindworkshop, final]{neurips_2025}

\usepackage[utf8]{inputenc} 
\usepackage[T1]{fontenc}    
\usepackage{url}            
\usepackage{booktabs}       
\usepackage{amsfonts}       
\usepackage{nicefrac}       
\usepackage{microtype}      
\usepackage{xcolor}         
\usepackage{graphicx}

\usepackage[colorlinks=true,
    citecolor=blue,    
    linkcolor=red,     
    urlcolor=magenta   
]{hyperref}

\title{Faithful Summarization of Consumer Health Queries: A Cross-Lingual Framework with LLMs}
\workshoptitle{Muslims in ML}

\author{%
  Ajwad Abrar \thanks{These authors contributed equally to the work.} \\
  Islamic University of Technology\\
  Dhaka, Bangladesh \\
  \texttt{ajwadabrar@iut-dhaka.edu} \\
  \And
  Nafisa Tabassum Oeshy \footnotemark[1]\\
  Islamic University of Technology\\
  Dhaka, Bangladesh \\
  \texttt{nafisatabassum4@iut-dhaka.edu} \\
  \AND 
  Prianka Maheru \footnotemark[1] \\
  Islamic University of Technology\\
  Dhaka, Bangladesh \\
  \texttt{priankamaheru@iut-dhaka.edu} \\
  \And
  Farzana Tabassum \\
  Islamic University of Technology\\
  Dhaka, Bangladesh \\
  \texttt{farzana@iut-dhaka.edu} \\
  \And
  Tareque Mohmud Chowdhury \\
    Islamic University of Technology\\
  Dhaka, Bangladesh \\
  \texttt{tareque@iut-dhaka.edu} \\
}

\begin{document}

\maketitle

\begin{abstract}
Summarizing consumer health questions (CHQs) can ease communication in healthcare, but unfaithful summaries that misrepresent medical details pose serious risks. We propose a framework that combines TextRank-based sentence extraction and medical named entity recognition with large language models (LLMs) to enhance faithfulness in medical text summarization. In our experiments, we fine-tuned the LLaMA-2-7B model on the MeQSum (English) and BanglaCHQ-Summ (Bangla) datasets, achieving consistent improvements across quality (ROUGE, BERTScore, readability) and faithfulness (SummaC, AlignScore) metrics, and outperforming zero-shot baselines and prior systems. Human evaluation further shows that over 80\% of generated summaries preserve critical medical information. These results highlight faithfulness as an essential dimension for reliable medical summarization and demonstrate the potential of our approach for safer deployment of LLMs in healthcare contexts.
\end{abstract}

\section{Introduction}

The rapid growth of online health consultations, especially consumer health questions (CHQs), has created both opportunities and challenges for healthcare delivery. These platforms, accelerated by the pandemic, now serve as vital sources of medical information and support. However, the large volume of verbose and sometimes redundant patient queries places a significant burden on healthcare professionals, who must spend time identifying the core concern before providing an appropriate response. Automatic medical text summarization has emerged as a potential solution to this problem by condensing lengthy questions into concise, focused forms.  

Traditional approaches to text summarization are primarily evaluated on \textit{general quality} metrics such as ROUGE or BERTScore, which measure lexical or semantic similarity to reference summaries. While useful, these metrics often fail to capture \textit{faithfulness}---the factual consistency of a summary with its source. Prior studies show that abstractive models frequently produce intrinsic errors (e.g., misrepresenting entities or relationships) and extrinsic errors (e.g., introducing unsupported facts) \citep{maynez2020faithfulness, huang2021factual}. In the medical domain, even minor distortions can mislead professionals or patients, posing direct risks to health outcomes. Thus, ensuring faithful summaries is essential for practical deployment in healthcare contexts.  

Despite the progress of large language models (LLMs), current systems still struggle to balance fluency, conciseness, and factual consistency in medical summarization. Faithfulness remains underexplored compared to readability or general accuracy, and most methods do not explicitly address it. This gap motivates the need for specialized frameworks that preserve the integrity of medical information while still providing concise and accessible summaries.  

To address this, we propose a novel framework that combines \textbf{TextRank-based sentence extraction} with \textbf{medical named entity recognition (NER)} to guide LLMs in generating summaries that are both accurate and faithful. We evaluate the framework on English (MeQSum) and Bangla (BanglaCHQ-Summ) datasets using LLaMA-2-7B fine-tuned with low-rank adaptation. Our main contributions are:  

\begin{itemize}
    \item We integrate extractive and abstractive methods to improve both informativeness and reliability.  
    \item We incorporate medical NER to ensure critical entities are preserved in summaries.  
    \item We provide the first cross-lingual evaluation of faithfulness in medical CHQ summarization, demonstrating improvements over zero-shot baselines and prior state-of-the-art systems.  
\end{itemize}

\section{Related Work}
Text summarization is a well-established task in NLP that aims to condense lengthy documents while preserving salient content \citep{el2021automatic}. Techniques generally fall into extractive methods, which select important sentences, and abstractive methods, which generate new phrasing \citep{rush2017neural}. While both have advanced with neural models and pretrained transformers, applying them to biomedical text is especially challenging due to domain-specific terminology, the complexity of clinical narratives, and the high stakes of factual accuracy \citep{afantenos2005summarization, morozovskii2023rare}.  

Medical summarization has been studied for clinical notes, conversations, and consumer health questions (CHQs). The MeQSum dataset \citep{ben-abacha-demner-fushman-2019-summarization} introduced 1,000 annotated CHQs and spurred research into neural abstractive approaches. More recently, BanglaCHQ-Summ \citep{khan-etal-2023-banglachq} extended this task to Bangla, highlighting cross-lingual gaps. Despite promising results with transformer-based models \citep{michalopoulos-etal-2022-medicalsum}, maintaining reliability across languages and medical domains remains an open challenge.  

A key limitation of existing work lies in \textit{faithfulness}—the factual consistency of generated summaries. Prior studies show abstractive models often introduce intrinsic errors (contradictions) or extrinsic errors (unsupported additions) \citep{maynez2020faithfulness, huang2021factual}. Recent solutions include constrained decoding \citep{mao2020constrained}, fact-checking modules \citep{kryscinski-etal-2020-evaluating}, and specialized evaluation metrics such as SummaC and AlignScore \citep{laban2021summac, zha2023alignscore}. However, ensuring factual alignment for consumer health queries, where even minor distortions can mislead patients or clinicians, remains underexplored. This gap motivates our focus on frameworks that explicitly preserve medical entities and source-grounded content.

\section{Proposed Methodology} \label{chapter:proposed}

\begin{figure}[t]
    \centering
    \includegraphics[width=\linewidth]{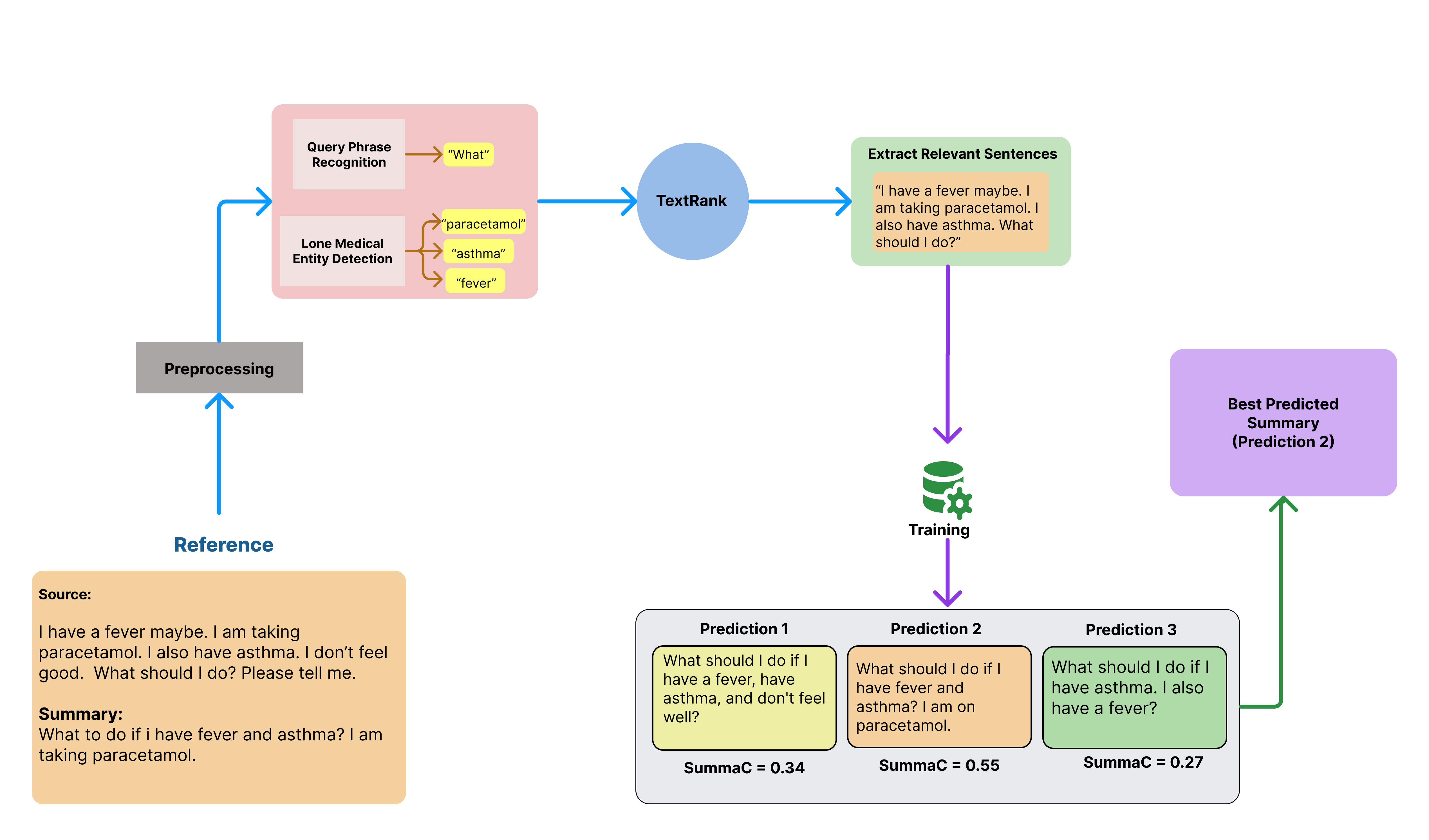}
    \caption{Proposed framework: TextRank extracts relevant sentences containing medical entities, which are used to fine-tune the LLM. The final summary is selected to maximize both accuracy and faithfulness.}
    \label{fig:methodology}
\end{figure}

We propose a framework to improve the faithfulness of medical text summarization, with a focus on Consumer Health Questions (CHQs) in English and Bangla. The framework combines \textit{Medical Named Entity Recognition (NER)} and the \textit{TextRank} algorithm for extractive sentence selection, followed by fine-tuning a Large Language Model (LLM) to generate accurate and reliable summaries. An overview is shown in \autoref{fig:methodology}.  

\subsection{Datasets}
We use two benchmark datasets. For English, the \textbf{MeQSum} dataset \citep{ben-abacha-demner-fushman-2019-summarization} contains 1,000 consumer health questions with expert-validated summaries, ensuring high-quality references. For Bangla, we use \textbf{BanglaCHQ-Summ} \citep{khan-etal-2023-banglachq}, which includes 2,350 annotated pairs of questions and summaries, representing the first large-scale resource for Bangla CHQ summarization. Together, these datasets enable evaluation across both high-resource and low-resource settings.  

\subsection{Preprocessing and Relevant Sentence Extraction}
Preprocessing standardizes the datasets into \textit{question} and \textit{summary} fields, while identifying overlapping medical entities and negation terms to ensure critical information is retained. To reduce noise, we apply the \textbf{TextRank} algorithm \citep{mihalcea2004textrank} to extract sentences containing medical entities and query-related words. This guarantees that summaries remain faithful to medically important content before abstractive generation by the LLM.  

\subsection{Evaluation Metrics}
We assess performance using both general quality and faithfulness metrics. General quality is measured by ROUGE-1/2/L and BERTScore, which capture lexical and semantic overlap. Faithfulness is measured by SummaC \citep{laban2021summac} and AlignScore \citep{zha2023alignscore}, which evaluate factual consistency, along with Flesch Reading Ease (FRE) for readability. This combination ensures that generated summaries are not only fluent but also factually aligned with the source.

\begin{table}[t]
\centering
\resizebox{\linewidth}{!}{
\begin{tabular}{lccccccc}
\toprule
\textbf{Setting (MeQSum)} & \textbf{R1} & \textbf{R2} & \textbf{RL} & \textbf{BERT} & \textbf{Read.} & \textbf{SummaC} & \textbf{AlignScore} \\ 
\midrule
Zero-shot (no FT)     & 21.97 & 6.48  & 19.98 & 0.60 & 65.16 & 0.28 & 21.80 \\
FT (no TR)            & 44.23 & 27.36 & 41.55 & 0.71 & 70.21 & 0.31 & 38.45 \\
FT + TR               & 47.07 & 29.44 & 44.08 & 0.72 & 70.69 & 0.37 & 45.65 \\
Best-of-3 (R1 select) & \textbf{50.50} & \textbf{34.38} & \textbf{47.74} & \textbf{0.74} & \textbf{71.56} & 0.40 & 39.24 \\
Best-of-3 (SummaC)    & 48.27 & 31.38 & 45.34 & 0.73 & 71.56 & \textbf{0.57} & \textbf{45.91} \\
\bottomrule
\end{tabular}}
\caption{MeQSum results across settings. Best-of-3 improves both quality and faithfulness; SummaC-based selection yields the highest factual consistency and strong alignment.}
\label{tab:meqsum_summary}
\end{table}

\section{Results and Discussion}
\label{sec:results}

\subsection{Performance Analysis (MeQSum)}
We evaluate LLaMA-2-7B in four settings: zero-shot, fine-tuning (FT) without TextRank, FT with TextRank-selected sentences (FT+TR), and best-of-3 selection using either ROUGE-1 (R1) or SummaC as the selector. Zero-shot underperforms, as expected for task-specific summarization \citep{abbasiantaeb2024let}. Fine-tuning substantially boosts both general quality and faithfulness, and adding TextRank further improves alignment with source content. Selecting the best of three candidates yields the strongest scores.

\vspace{2mm}
\noindent\textbf{Temperature.} We sweep $t\in\{0.1,0.3,0.5,0.7,0.9\}$ and observe a trade-off: lower $t$ favors ROUGE, higher $t$ favors SummaC. We adopt $t{=}0.7$ as a balanced choice, giving peak faithfulness with competitive ROUGE \citep{yucan}.

\noindent\textbf{Comparison to prior work.} Our approach surpasses strong baselines on MeQSum, particularly in R1/RL and readability, while achieving competitive or superior faithfulness.

\begin{table}[t]
\centering
\resizebox{\linewidth}{!}{
\begin{tabular}{lccccccc}
\toprule
\textbf{Model} & \textbf{R1} & \textbf{R2} & \textbf{RL} & \textbf{BERT} & \textbf{Read.} & \textbf{SummaC} & \textbf{AlignScore} \\
\midrule
Mixtral-8x7B-Inst. \citep{dada2024clue} & 32.47 & 36.38 & 16.86 & 0.72 & -- & -- & -- \\
BioBART + FaMeSumm \citep{zhang2023famesumm} & 31.76 & 11.71 & 29.64 & 0.74 & -- & 0.46 & -- \\
\textbf{Ours (Best-of-3)} & \textbf{50.50} & \textbf{34.38} & \textbf{47.74} & \textbf{0.74} & \textbf{71.56} & \textbf{0.57} & \textbf{0.46} \\
\bottomrule
\end{tabular}}
\caption{State-of-the-art comparison on MeQSum. Our method achieves the best performance across most metrics, with strong factual consistency (SummaC) and alignment (AlignScore).}
\label{tab:sota}
\end{table}

\subsection{Performance Analysis (BanglaCHQ-Summ)}
We replicate the evaluation on BanglaCHQ-Summ to assess cross-lingual robustness. Zero-shot performance is weak, reflecting linguistic/domain gaps. Fine-tuning improves all metrics, and FT+TR yields further gains. Best-of-3 selection again provides the strongest outcomes; SummaC-based selection maximizes faithfulness.

\begin{table}[h]
\centering
\resizebox{\linewidth}{!}{
\begin{tabular}{lcccc}
\toprule
\textbf{Setting (Bangla)} & \textbf{R1} & \textbf{R2} & \textbf{RL} & \textbf{BERT} \ \ \textbf{SummaC} \\
\midrule
Zero-shot (no FT)         & 19.10 & 8.21  & 18.97 & 0.62 \ \ 0.22 \\
FT (no TR)                & 28.24 & 14.22 & 24.54 & 0.71 \ \ 0.26 \\
FT + TR                   & 30.71 & 15.71 & 28.95 & 0.74 \ \ 0.28 \\
Best-of-3 (R1 select)     & \textbf{32.35} & \textbf{16.32} & \textbf{29.09} & \textbf{0.76} \ \ 0.29 \\
Best-of-3 (SummaC select) & 30.92 & 15.74 & 27.35 & 0.73 \ \ \textbf{0.32} \\
\bottomrule
\end{tabular}}
\caption{BanglaCHQ-Summ results. Best-of-3 improves quality (R1/RL/BERT) and faithfulness (SummaC).}
\label{tab:bangla_summary}
\end{table}

\subsection{Human Evaluation}
To validate the reliability of generated summaries, we conducted a human evaluation on the MeQSum dataset with the help of a medical doctor. The evaluation focused on two questions: (1) whether the summary retained all critical information from the source text, and (2) whether it was factually consistent. A binary (yes/no) judgment was provided for each case. A summary was considered faithful only if both conditions were met. Results showed that 82\% of the summaries satisfied these criteria, demonstrating strong factual alignment. 

\section{Conclusion}
We proposed a framework that combines TextRank-based extraction, medical NER, and fine-tuned LLaMA-2-7B to enhance the faithfulness of medical text summarization. Experiments on English (MeQSum) and Bangla (BanglaCHQ-Summ) datasets show consistent gains over zero-shot and prior systems in both quality and factual consistency. Human evaluation further confirmed the reliability of our approach, with over 80\% of summaries judged as faithful. A limitation of this study is that experiments were conducted on a single LLM (LLaMA-2-7B) and two languages, leaving scope for future work to explore multiple LLMs and broader multilingual settings. Future directions include adapting the framework to few-shot and multilingual settings, extending to other sensitive domains (e.g., legal and financial texts), and incorporating expert feedback for improved robustness and practical integration into clinical workflows.


\bibliographystyle{plainnat}  
\bibliography{citations}       

\medskip

\small

\end{document}